\documentclass[final]{cvpr}

\usepackage{times}
\usepackage{epsfig}
\usepackage{graphicx}
\usepackage{amsmath}
\usepackage{amssymb}
\usepackage{algorithm}
\usepackage{algorithmic}
\usepackage{multirow}
\usepackage{float}
\usepackage{color}

\usepackage[pagebackref=true,breaklinks=true,colorlinks,bookmarks=false]{hyperref}

\def\cvprPaperID{3296} 
\def\confYear{CVPR 2021}

\begin{document}
	
	\title{A Backbone Replaceable Fine-tuning Framework for Stable Face Alignment}
	
	\author{
		Xu Sun$^{1,2}$, Zhenfeng Fan $^{1}$, Zihao Zhang $^{1,2}$, Yingjie Guo$^{1}$, Shihong Xia$^{1,2}$\\
		{\tt\small $[$sunxu, fanzhenfeng, zhangzihao, guoyingjie, xsh$]$@ict.ac.cn}\\
		{\footnotesize 
			$^{1}$Institute of Computing Technology, Chinese Academy of Sciences, 
			$^{2}$University of Chinese Academy of Sciences}
	}
	
	\maketitle

	\begin{abstract}
		Heatmap regression based face alignment has achieved prominent performance on static images. However, the stability and accuracy are remarkably discounted when applying the existing methods on dynamic videos. We attribute the degradation to random noise and motion blur, which are common in videos. The temporal information is critical to address this issue yet not fully considered in the existing works. In this paper, we visit the video-oriented face alignment problem in two perspectives: detection accuracy prefers lower error for a single frame, and detection consistency forces better stability between adjacent frames. On this basis, we propose a Jitter loss function that leverages temporal information to suppress inaccurate as well as jittered landmarks. The Jitter loss is involved in a novel framework with a fine-tuning ConvLSTM structure over a backbone replaceable network. We further demonstrate that accurate and stable landmarks are associated with different regions with overlaps in a canonical coordinate, based on which the proposed Jitter loss facilitates the optimization process during training. The proposed framework achieves at least 40\% improvement on stability evaluation metrics while enhancing detection accuracy versus state-of-the-art methods. Generally, it can swiftly convert a landmark detector for facial images to a better-performing one for videos without retraining the entire model.
	\end{abstract}
	
	\section{Introduction}
	\emph{Facial landmark detection} (\textit{a.k.a.} face alignment) is an active branch of research in the field of computer vision.
	It has been widely used in numerous facial related applications such as face recognition~\cite{ChangLCW20, deng2019arcface, DuongTLQBR20, GuoZZCLL20, MasiCCHKKLRWHAM19, WangWSC17, ZhangDWHLZW20} and 3D face reconstruction~\cite{LeeL20a, LinYSZ20, ZZZ018, zhou2019dense, ZhuWCVW20}.
	
	From the traditional feature and shape regression based schemes to the recently extensive used convolutional neural network (CNN) based heatmap regression strategies, facial landmark detection algorithms have undergone tremendous development over the past few decades.
	In view of the powerful image feature extraction capabilities of CNN, modern face alignment approaches have widely employed CNN as backbone.
	Among them, the early methods are designed to directly regress the coordinates of landmarks~\cite{feng2018wing, miao2018direct, PengZYM18, wu2017leveraging, TuzelMT16}, which require adding a fully connected layer at the end of network.
	However, regressing coordinates directly from images is an extremely nonlinear process, which leads to inferior generalization of trained models. 
	This usually raises difficulty to solve problem in complex situations such as large occlusion and bad illumination.
	Therefore, coordinate regression is rapidly replaced by \emph{heatmap regression}.
	Heatmap based approaches~\cite{BrowatzkiW20, bulat2017far, ChandranBGB20, 0004MM00CKL020, TangPLM20} predict probability map for each landmark, which allows its structure to be fully convolutional.
	Furthermore, heatmap has a better discrimination between foreground and background as an intermediate result.
	These properties make predicted landmarks via heatmaps more robust than coordinates regression methods~\cite{BulatT20, JourablooY0R17, merget2018robust, RenCWS16, valle2018deeply}.

	\begin{figure}[t!]
		\begin{center}
			\includegraphics[width=0.490\linewidth]{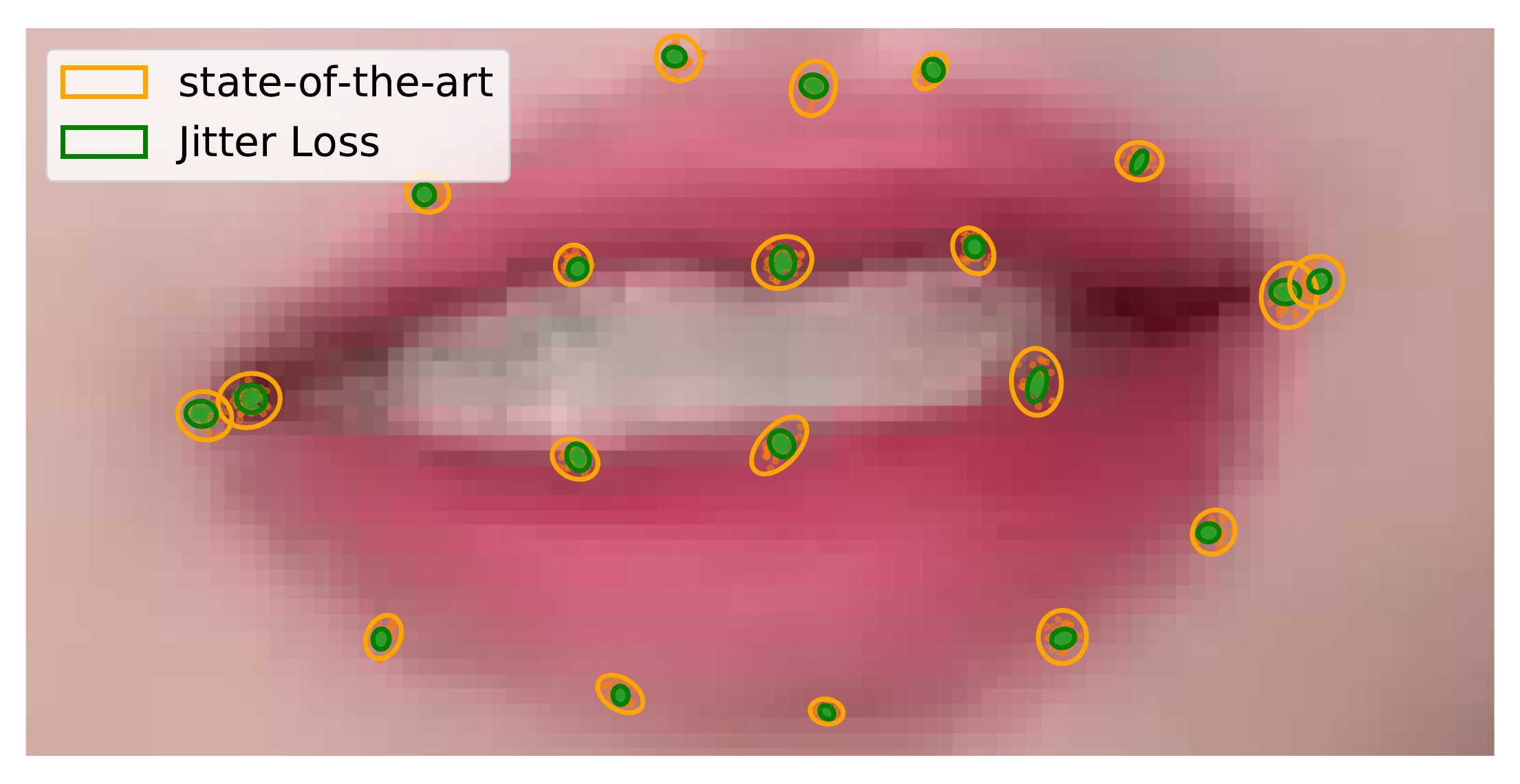}
			\includegraphics[width=0.500\linewidth]{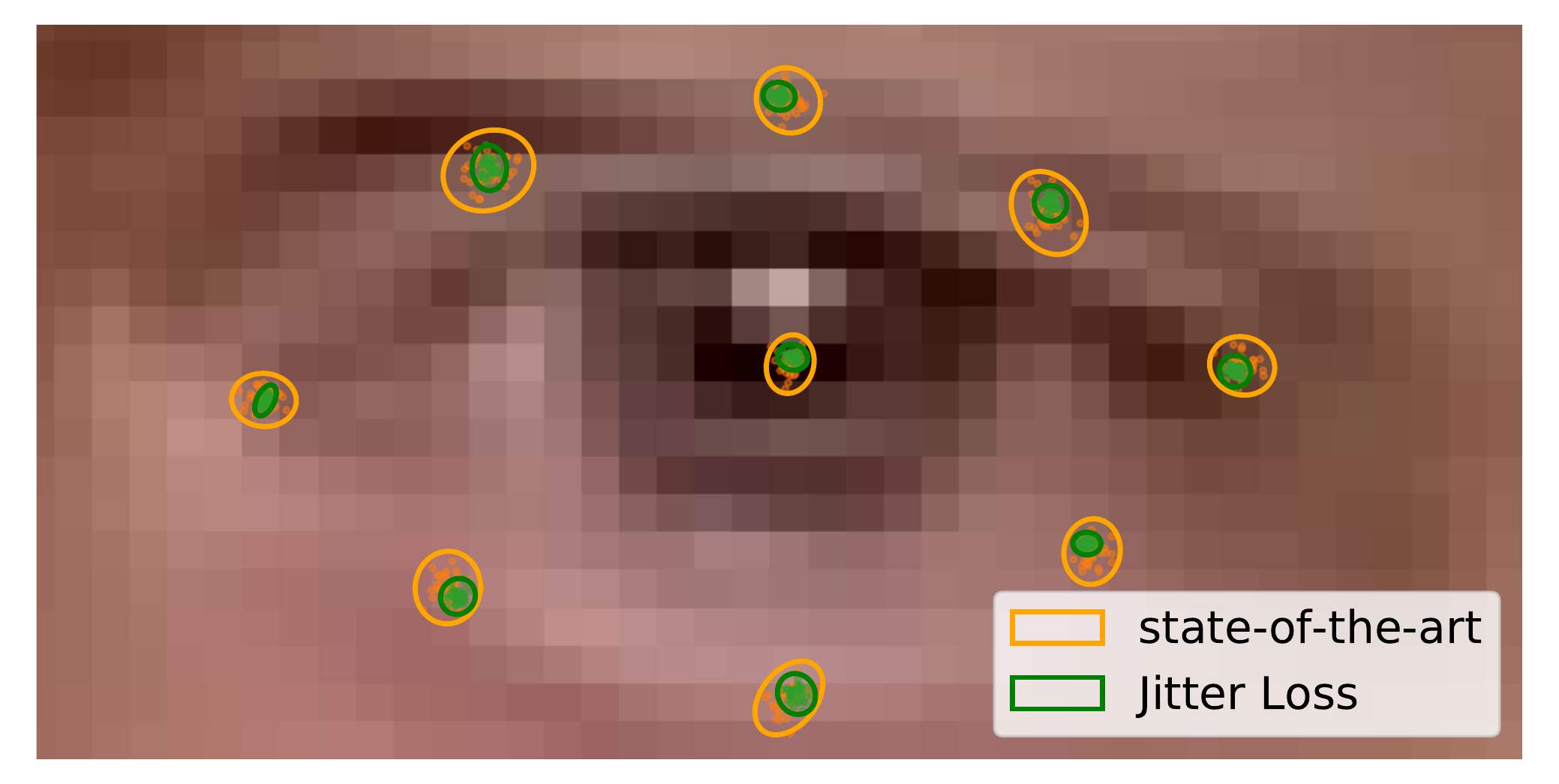}
		\end{center}
		\vspace*{-0.1in}
		\caption{
			Comparison of stability for landmarks generated from $30$ consecutive frames.
			The orange ellipse indicates predictions within $95\%$ confidence via a state-of-the-art method~\cite{wang2019adaptive},
			while the green one denotes the results obtained by the fine-tuned model in this work.
			The proposed method
			significantly shrink the distribution of predicted landmarks with improved stability.
		}
		\vspace*{-0.1in}
		\label{fig:Logo}
	\end{figure}
	
	Heatmap regression based facial landmark detectors have achieved prominent performance at public datasets of static images.
	However, some issues remain to be solved when applying them to real-life applications.
	Among them, predicted landmarks jitter in videos is the most urgent problem.
	Notable jitter of predict landmarks still exists when applying state-of-the-art (SOTA) methods to sequential images or successive frames. Figure~\ref{fig:Logo} shows an example.
	
	Since numerous face recognition methods rely on accurate landmarks for face alignment, landmark errors may hinder the performance of these methods.
	Moreover, in face animation~\cite{CaoWWS14, wang2016realtime} and reenactment~\cite{HuangYW20, YaoYSZ20} methods, 2D landmarks are used as anchors to deform 3D face meshes toward realistic facial performances, thus \emph{temporal jitter} of detected 2D landmarks in video will propagate to 3D mesh, which could generate perceptually jarring results.

	The cause of jitter mainly lies in the following aspects:
	Firstly, \emph{random noise} and \emph{motion blurs} are common in videos captured via non-professional camera.
	Unfortunately, heatmap regression is extremely sensitive to abnormal data.
	The most common network structures are also designed for static images without considering temporal information of videos. 
	In such cases, landmarks would be jittery even if the face is not moving.
	Secondly, existing loss functions cannot effectively employ \emph{temporal information}, which brings about their inability to
	impose adequate penalties on the jittering predictions during training.
	Finally, existing post-processing methods to predict landmarks from heatmaps are sensitive to tiny errors on background pixels, which also leads to instable predictions.
	The above reasons explain why the performance of existing algorithms are degraded while testing in videos.
	
	In this paper, we propose a basic network structure, a loss function, and a post-processing method for video-based face alignment. We further propose two evaluation metrics which are suitable for videos to judge jitter phenomenon.
	In summary, the main contributions of this work are three-fold:
	
	\begin{itemize}
		\item We design a brand new backbone replaceable framework for training video-oriented facial landmark detector, which can quickly adapt a image-oriented detector to a video-oriented one with lifted performance, especially stability.
		\item We propose a novel Jitter loss, which incorporates temporal information to impose strong penalties on predicted landmarks that jitter around the ground truth.
		\item We introduce two evaluation metrics for landmark stability measurement, and the baselines on public video datasets are given to facilitate future research.
	\end{itemize}
	
	\section{Related Work}
	\paragraph{Video-based Facial Landmark Detection.}
	Image-based facial landmark detectors have achieved splendid performance on static pictures.
	However, when these methods are applied to videos or sequential images, the precision and stability is remarkably discounted.
	Hence, a number of video-based face alignment methods have been proposed.
	Pure temporal tracking~\cite{hare2015struck, shen2015first} is a common method for video-based detectors.
	But it usually suffers from tracker drift, which denotes tracking errors in the current frame may affect subsequent video frames.
	Therefore, some hybrid methods~\cite{LiuLFZ18, peng2016recurrent, trigeorgis2016mnemonic, YinWCCL20, ZhuLLDT20} are proposed, which jointly utilize tracking-by-detection and temporal information in a single framework to predict more stable facial landmarks.
	These methods utilize recurrent neural networks (RNN) to encode the temporal information across consecutive frames.
	Based on RNN and incremental learning, personalized face alignment method~\cite{PengZYM18} has been proposed to address the issue of sensitivity to initializations, which adopts multiple initial shapes to regress the facial landmarks within one frame.
	To step further, the SBR method~\cite{dong2018supervision} uses the result of the previous frame to assist the detection of landmarks, but it is sensitive to digital noise.
	The FHR approach~\cite{tai2019towards}, which is considered better robustness to noise, leverages 2D Gaussian generation priors to accurately estimate the fraction part of coordinates, however, it does not work well in low-resolution videos.
	Since these methods for stabilization are all end-to-end, the whole network needs to be retrained when the test data change from images to videos.
	In this work, we propose a novel fine-tuning framework to quickly convert existing \emph{image-oriented} detectors to \emph{video-oriented} ones.
	The proposed framework requires minor effort for training and boosts the performance greatly.
	
	\paragraph{Loss Function for Heatmap Regression.}
	Mean square error (MSE, \textit{a.k.a.} $\mathcal {L}_2$) is the most common loss function for regression.
	However, the gradient of MSE is linear, so at the later stage of training, it is difficult to impose strong constraints on predicted landmarks which are close to the ground truth.
	The over-sensitivity of MSE to outliers also leads to gradient explosions.
	To solve the above issues, Wing loss~\cite{feng2018wing} provides a constant gradient for large errors to avoid gradient explosion, meanwhile, it amplifies the gradient near the origin through a piecewise exponential function.
	However, the gradient of Wing loss is discontinuous around the origin, which incurs a significant increase in training time.
	In addition, Wing loss is originally designed for coordinate regression rather than heatmap regression due to its high sensitivity to tiny errors on background pixels.
	Adaptive Wing (AWing) loss~\cite{wang2019adaptive}, as an improvement of Wing loss, overcomes some of the difficulties by adapting its shape to different heatmaps.
	Unfortunately, all the above losses are designed for still images without mechanism for stable landmarks across videos sequences.
	In this study, we propose a novel Jitter loss to leverage temporal information for robust prediction of landmarks.
	
	\begin{figure*}[t!]
		\begin{center}
			\includegraphics[width=1\linewidth]{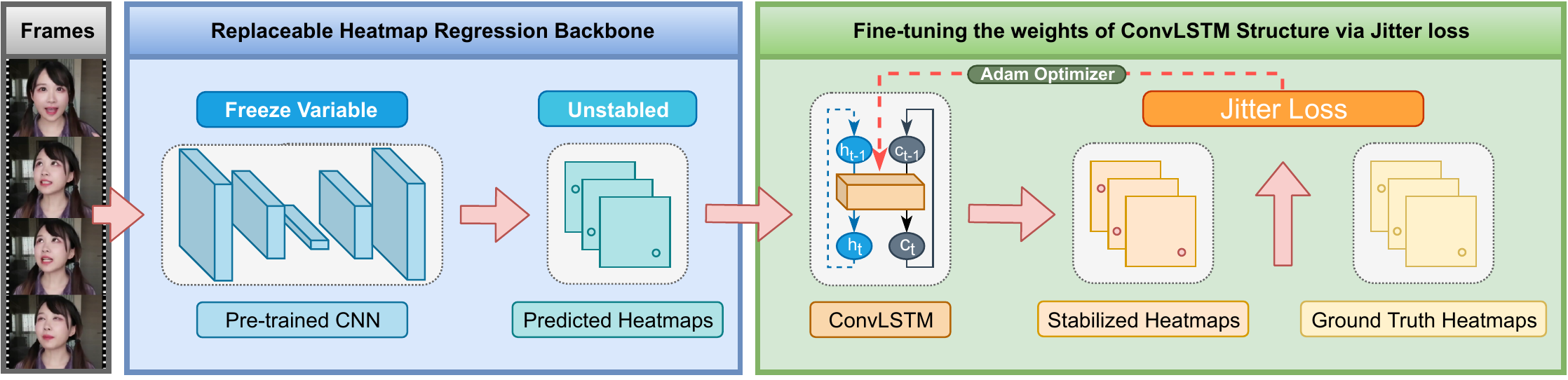}
		\end{center}
		\vspace*{-0.1in}
		\caption{
			An overview of our two-stage network.
			The heatmaps calculated from backbone heatmap regression on the first stage are fed into the input of the second stage to generate stabilized predictions, where $h_t$ and $c_t$ are used as hidden states to record temporal information. 
			Stabilized predictions of heatmaps and the ground truth are applied to Jitter loss for fine-tuning, and the optimizer updates the weights of ConvLSTM based on the calculated loss value.
		}
		\label{fig:pipeline}
		\vspace*{-0.1in}
	\end{figure*}

	\section{Proposed Method}
	The proposed method incorporates a backbone replaceable network as the basic facial landmark detector. The detector is either a publicly available model or a re-trained one with certain datasets. This provides flexibility for users to customize the backbone network to specific usages with minor effort. In this section, we describe the basic framework, the loss function, and the post-processing procedure which cover the details of the proposed method.
	
	\subsection{Fine-tuning Framework}
	Modern cameras commonly generate an image via an ensemble of many captured images when users hold the camera still. This reduces the noise and blurring effect of the resulted image. However, we notice that the videos are often masked by trenmendous noise and blurring effect in a departure from general images. Thus there commonly exists false detections and jittery predictions when applying a detector designed for images to video frames. We propose to leverage the temporal information, in particular \emph{the coherence of adjacent frames} in video sequence into our designed framework
	for more robust landmark detection.
	
	The proposed two-stage network is shown in Figure~\ref{fig:pipeline}. 
	On the first stage, a trained backbone network is employed as feature extractor to generate feature maps. This network can be replaced by any compatible structures that support heatmap regression such as the HRNet~\cite{sun2019high}, the CPM~\cite{WeiRKS16}, and the HourGlass~\cite{newell2016stacked}.
	On the second stage, we employ the \textbf{ConvLSTM}~\cite{xingjian2015convolutional} structure into our 
	framework
	to make full use of the temporal information to generate \emph{stable} facial landmarks.
	This gives an easy access to generalize a landmark detector for facial images to that for video sequences.
	
	Given the heatmaps $o_t$ generated by an original network in frame $I_t$ with the hidden states $h_{t-1}$ and $c_{t-1}$ from the previous processed frame, the stabilized heatmap $s_t$ on the current frame with reserved hidden states $h_{t}$ and $c_{t}$ yielded by ConvLSTM can be defined as follows:
	\begin{equation}
	h_t, c_t, s_t = ConvLSTM(h_{t-1}, c_{t-1}, o_t).
	\end{equation}
	The heatmap $s_t$ and its corresponding ground truth $\hat{s}_t$ are fed into a Jitter loss, which will be described afterward. Then the optimizer updates the weights of ConvLSTM iteratively.
	By this way, we make the loss differentiable to the features from both the current and the previous frames.
	We also include hidden states generated from previous frames to infer more stable results.
	This has advantage to be more tolerant to prediction errors for individual frames and thus avoids jitter predictions.
	
	\subsection{Jitter Loss}
	Existing loss functions for heatmap regression require the calculations of pixel-wise errors between the prediction and the ground truth. This metric is proved to work well on static images in the literature. However, the loss values as functions of the difference between pixels on a single frame cannot adequately suppress the negtive effect that the predicted landmarks jitter around the target point. As a result, the predictions of video-based face alignment models trained via such loss functions usually suffer from remarkable jittering. To this end, we propose a \textbf{Jitter loss} function to capture large jitter and impose strong penalty on it during training, which is defined as follows:
	\begin{equation}\label{e2}
	\mathcal{L}_{Jitter}(s_{t},\hat{s}_{t})=\Psi\cdot\mathcal{L}_{pixel}(s_{t},\hat{s}_{t}),
	\end{equation}
	where $s_{t}$ and $\hat{s}_{t}$ are the predicted heatmap and the ground truth 
	for the $t^{th}$ frame, respectively. This loss includes a \emph{modulation} term over an original \emph{pixel} loss, which takes into consideration not only the prediction error of each frame but also the jittering effect for adjacent frames. 
	
	The modulation term $\Psi$ is defined as follows:
	\begin{equation}\label{e3}
	\Psi=\min\{\frac{{\left\| {({u_t} - {{\hat u}_t}) - ({u_{t - 1}} - {{\hat u}_{t - 1}})} \right\|}_2}{{\left\| {{{\hat u}_t} - {{\hat u}_{t - 1}}} \right\|_2 + \xi }},\Theta\},
	\end{equation}
	where $u_t$ and ${\hat u}_t$ denote the predicted landmark and the ground truth of the $t^{th}$ frame, respectively. This function is upper-bounded by a hyperparameter $\Theta$ to avoid gradient explosions and its denominator includes a regularization parameter $\xi$ to avoid singular values during training. 
	
\begin{figure}[H]
		\vspace*{-0.1in}
		\centering
		\includegraphics[width=0.685\linewidth]{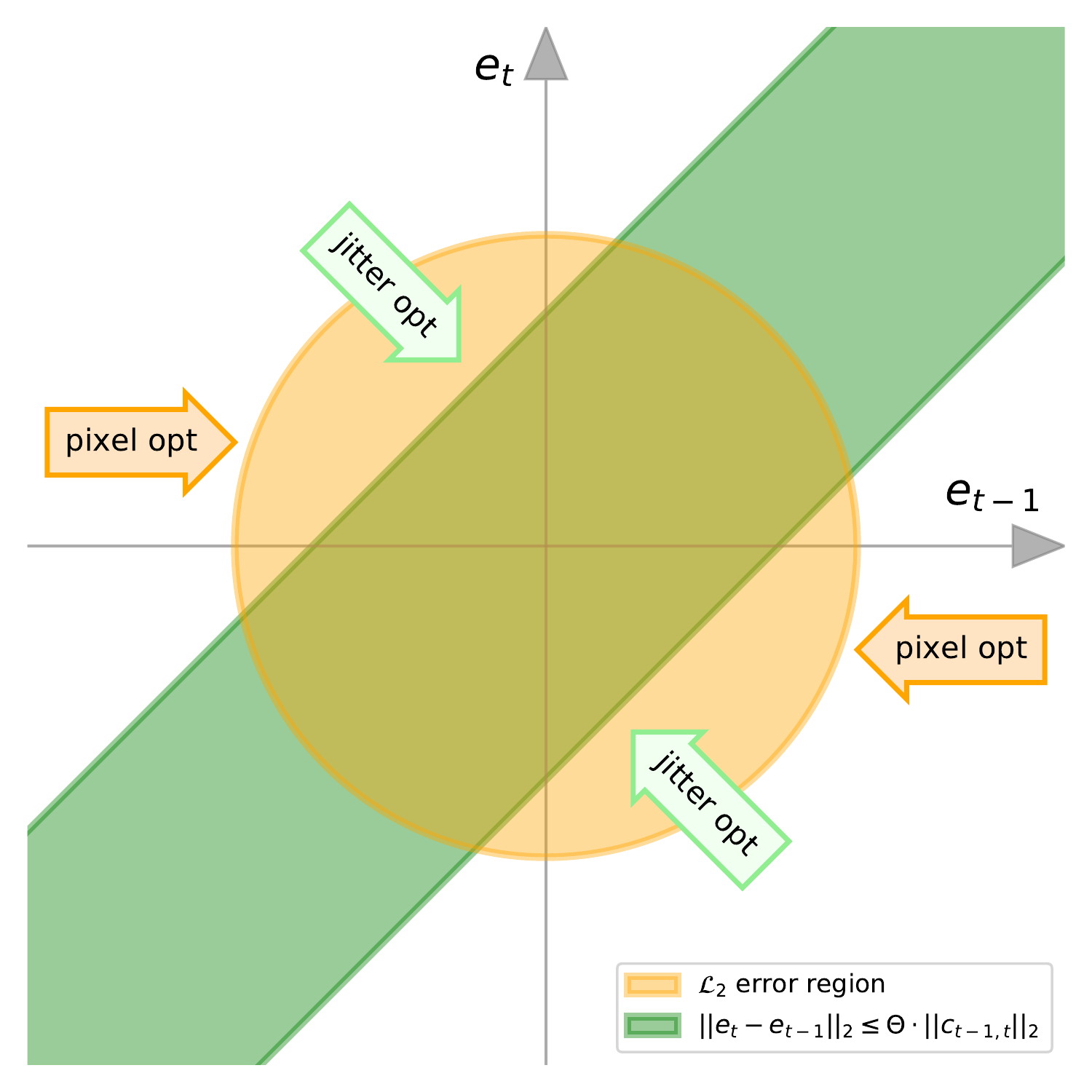}
		\caption{Illustrations of $\mathcal{L}_2$ error and jitter, both of which are our goals for training a robust landmark detector. }
		\label{fig:Jitter}
	\end{figure}

	Let $e_t={u_t} - {{\hat u}_t}$ and $e_{t-1}={u_{t - 1}} - {{\hat u}_{t - 1}}$ be the landmark prediction errors of two adjacent frames, respectively, and let $c_{t-1,t}={{\hat u}_t} - {{\hat u}_{t - 1}}$ be the ground truth landmark offset between the two frames. Then Eq.~\ref{e3} is equivalent to
	\begin{equation}\label{e4}
	\Psi=\min\{\frac{{\left\| {e_t - e_{t-1}} \right\|}_2}{{\left\| c_{t-1,t} \right\|_2 + \xi }},\Theta\}.
	\end{equation}
	This means that the Jitter loss in Eq.~\ref{e2} is \emph{linearly} dependent on the landmark inconsistency $e_t-e_{t-1}$ normalized by the ground truth landmark offset $c_{t-1,t}$ while thresholding by $\Theta$. We plot the relationship between them in Figure~\ref{fig:Jitter}. Note that we simplify the 2-D vectors into 1-D ones for clearer illustrations. 
	
	\textbf{Jitter criterion.} Then it comes up with the critical \emph{jitter criterion} according to Figure~\ref{fig:Jitter}:
	\begin{equation}\label{e5}
	{\left\| {{e_t} - {e_{t - 1}}} \right\|_2} > \Theta  \cdot \left\| {{c_{t - 1,t}}} \right\|_2,
	\end{equation}
	which is exactly in consistency with Eq.~\ref{e4}. The jitter phenomenon occurs when the inconsistency between predicted landmarks of two adjacent frames is larger than a certain threshold. We assign a large and fixed weight $\Theta$ to strengthen the Jitter loss. Otherwise, the Jitter loss is shrinked by a factor which is proportional to landmark inconsistency with respect to the ground truth landmark offset. This is in accordance with our physical insight on jittered landmarks. Note that the predict landmark location $u_t$ is obtained by mapping the predicted heatmap $s_t$ of the current frame via a post-processing algorithm as:
	\begin{equation}\label{e6}
	u_t = PostProcessing(s_t),
	\end{equation}
	of which the details are described in Section~\ref{sec:post}.
	
\begin{figure}[H]
		\vspace*{-0.1in}
		\begin{center}
			\includegraphics[width=\linewidth]{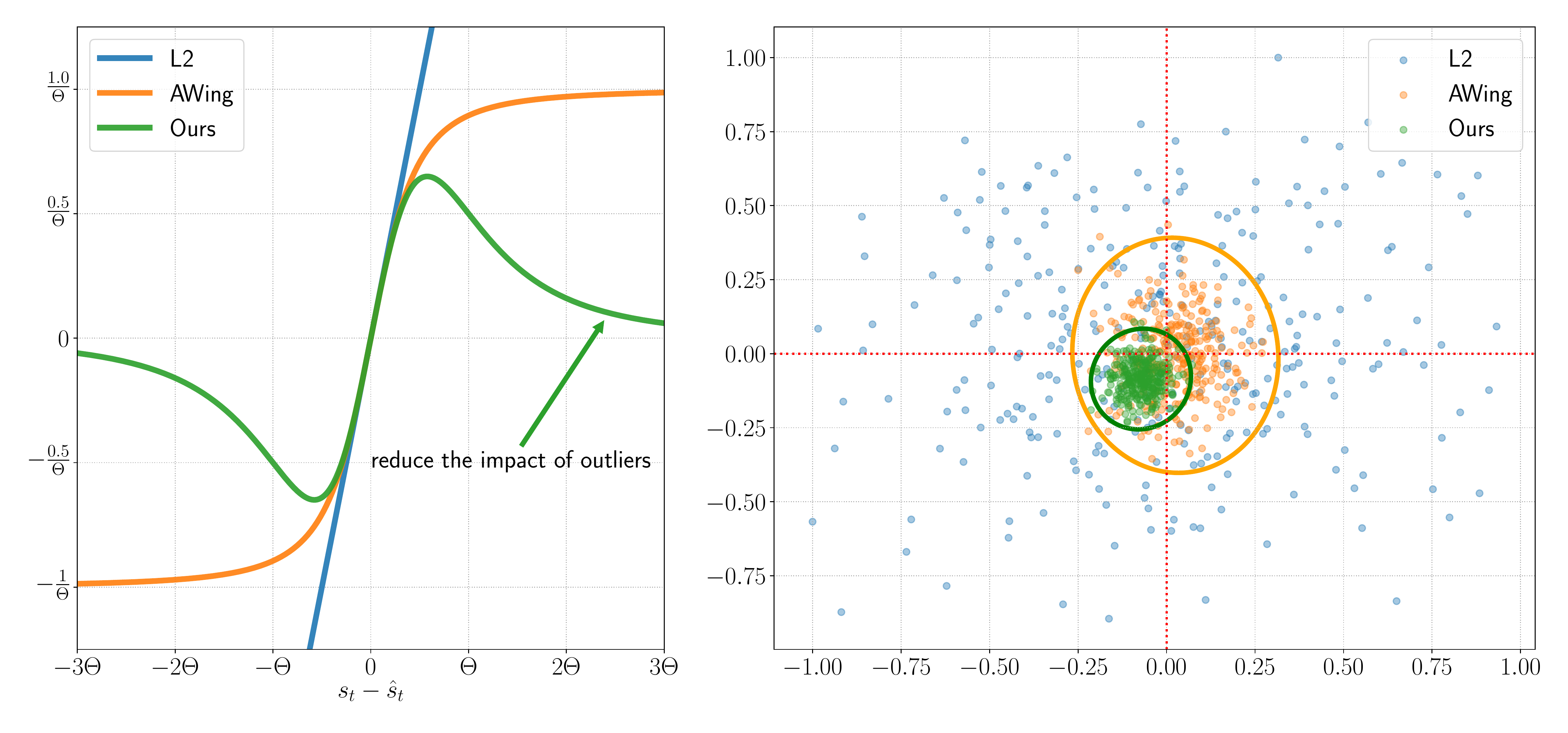}
		\end{center}
		\vspace*{-0.4cm}
		\caption{
			Comparison among $\mathcal{L}_2$, AWing loss, and Jitter loss.
			The gradient plot on the left shows that the backbone loss we choose for pixel error calculation can effectively penalize jittery predictions while reducing the impact of outliers.
			Normalized test results on the right prove that Jitter loss can significantly shrink distribution radius and slightly decrease the Normalized Mean Error at the same time.
		}
		\vspace*{-0.1in}
		\label{fig:CompareWithOthers}
	\end{figure}

	\textbf{Choice of pixel loss.} It is necessary to choose an appropriate pixel loss between $s_{t}$ and $\hat{s}_{t}$ in Eq.~\ref{e2}. Since our framework already pre-assumes a well-trained backbone network, \emph{fine-tuning} instead of training from scratch is required. More specifically, the proposed framework should focus on handling small errors through training the ConvLSTM in the latter stage while fixing the weights of the backbone network. We do not choose the $\mathcal{L}_2$ and smoothed $\mathcal{L}_1$ loss because their gradients are small around the origin.
	Wing and $\mathcal{L}_1$ loss alleviate the small-gradient problem, but they induce discontinuity on the origin which increases the difficulty for fine-tuning. AWing loss, as an improvement of Wing loss, can handle small error while ensuring continuous gradient. However, it is originally designed for handing outliers in a departure of our goal for handing jitters. Therefore, we choose the \emph{Geman-McClur} loss~\cite{barron2019general}, which has a reasonable gradient distribution as in Figure~\ref{fig:CompareWithOthers} (the left). In this study, the definition of the pixel loss is parameterized by the jitter threshold $\Theta$, as follows:
	\begin{equation}\label{e7}
	\mathcal{L}_{pixel}(s_{t},\hat{s}_{t})=
	\frac{(s_{t}-\hat{s}_{t})^2}
	{(s_{t}-\hat{s}_{t})^2+\Theta^2}.
	\end{equation}
	The gradient of the loss in Eq.~\ref{e7} reaches an extreme when $\left|s_{t}-\hat{s}_{t}\right|$ increases to $\Theta/2$ and then gradually decreases to zero, which enables the optimizer quite robust to outliers, thereby focusing on small errors caused by jittered landmarks. Figure~\ref{fig:CompareWithOthers} (the left) also includes the gradients of another two loss functions which are commonly used in the literature for comparison. Additionly, we conduct a normalized test for the error distributions of landmarks using the three different loss functions. The results are shown in Figure~\ref{fig:CompareWithOthers} (the right), where the distributions of landmark errors by the proposed Jitter loss are reduced significantly.
	
\begin{figure}[t!]
		\begin{center}
			\includegraphics[width=0.96\linewidth]{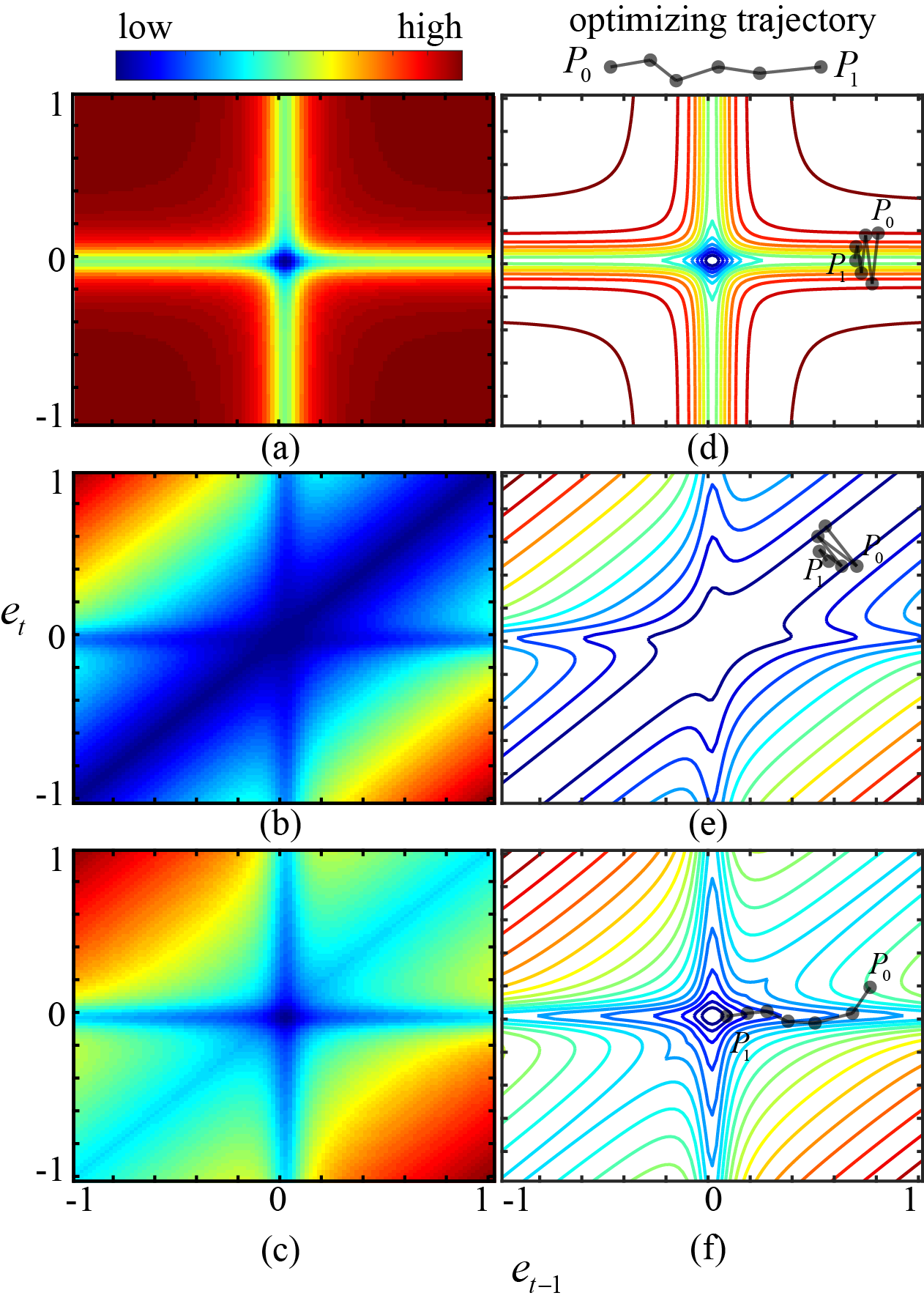}
		\end{center}
		\vspace*{-0.1in}
		\caption{
			Different loss functions with respect to prediction errors of two adjacent frames. The first column shows the loss values in jetmap and the second column shows the corresponding contour maps. From top to bottom: the pixel loss, the modulated pixel loss, and the proposed Jitter loss as an ensemble of them. Examples of optimizing trajectories (from $P_0$ to $P_1$) are plotted on the contour maps.
		}
		\label{fig:heatmap2D}
	\end{figure}

	\textbf{Effect of Jitter loss.} The above analysis has fully discussed the pixel loss we choose. We further elaborate the effect of Jitter loss with the modulation term $\Psi$. According to Eq.~\ref{e2} and Eq.~\ref{e3}, the Jitter loss modulated by $\Psi$ is a function of the regressed heatmaps for two adjacent frames.
	Since the unnormalized threshold $\Theta\cdot||c_{t-1,t}||_2$
	is dependent on individual training samples, we only illustate the mutual effects of all samples and take the landmark errors instead of the heatmaps for brevity. Therefore, the Jitter loss in Eq.~\ref{e2} can be considered as a combination of two terms, as follows:
	\begin{equation}\label{e8}
	\begin{aligned}
	&\mathcal{L}_{Jitter}(e_{t},e_{t-1})\\
	&=\lambda\cdot\{\mathcal{L}_{pixel}(e_{t})+\mathcal{L}_{pixel}(e_{t-1})\}\\
	&+{\left\| {{e_t} - {e_{t - 1}}} \right\|_2}\cdot\{\mathcal{L}_{pixel}(e_{t})+\mathcal{L}_{pixel}(e_{t-1})\},
	\end{aligned}
	\end{equation}
	where the weight $\lambda$ is determined by the trained results in progress adaptively. Figure~\ref{fig:heatmap2D} (a-c) includes the loss in Eq.~\ref{e8} and those contributed by the two individual terms on the right of Eq.~\ref{e8}, as functions to the landmark prediction errors of two adjacent frames. We also plot the corresponding contour maps for each of them in Figure~\ref{fig:heatmap2D} (d-e) to illustrate the gradient map. Note that all the values are normalized to the range $[-1,1]$. The prediction tends to oscillate around the axis $e_{t-1}=0$ or $e_t=0$ for a given initialization as shown in the optimizing trajectoty in Figure~\ref{fig:heatmap2D} (d), in case of which we only include the pixel loss (the first term of Eq.~\ref{e8}). When it applys to the modulated pixel loss (the second term of Eq.~\ref{e8}), the prediction tends to oscillate around the axis $e_{t}=e_{t-1}$ (see Figure~\ref{fig:heatmap2D} (e)). The two cases lead to either undesired jittering effects or large prediction errors. Our proposed Jitter loss (see Figure~\ref{fig:heatmap2D} (c,f)) overcomes the optimization difficulties as a combination of the two terms, which is the key to robust results for landmark predictions. 
	
	\begin{algorithm}[H]
		\caption{Probability Density Centralization}
		\label{alg:algorithm}
		\textbf{Input}: Heatmap: $\phi$\\
		\textbf{Parameter}: Threshold: $\Theta_{PDC}$, Height: $H$, Width: $W$\\
		\textbf{Output}: Coordinate: $[x, y]$
		\begin{algorithmic}[1] 
			\STATE Set all values less than $\Theta_{PDC}$ in $\phi$ to 0.
			\STATE $sum_{\phi} \leftarrow \sum \phi$
			
			\STATE $sum_{X} \leftarrow 0$
			\FOR{$i=0:W-1$}
			\STATE $sum_{X} += (i+1) * \sum_{h=0}^H \phi_{h, i}$
			\ENDFOR
			
			\STATE $sum_{Y} \leftarrow 0$
			\FOR{$i=0:H-1$}
			\STATE $sum_{Y} += (i+1) * \sum_{w=0}^W \phi_{i, w}$
			\ENDFOR
			\STATE $x \leftarrow {sum_{X}}/{sum_{\phi}}$
			\STATE $y \leftarrow {sum_{Y}}/{sum_{\phi}}$
			\STATE \textbf{return} $[x, y]$
		\end{algorithmic}
	\end{algorithm}

		\begin{table*}
			\centering
				\resizebox{\textwidth}{!}{
					\begin{tabular}{|c|p{0.8cm}p{0.8cm}p{0.8cm}p{1cm}|p{0.8cm}p{1cm}|p{0.8cm}p{1cm}|p{0.9cm}p{1cm}p{1cm}|}
						\hline
						\multirow{2}{*}{Method} & \multicolumn{4}{c|}{300W(NME)} & \multicolumn{2}{c|}{AFLW(NME)} & \multicolumn{2}{c|}{COFW} & \multicolumn{3}{c|}{WFLW} \\
						& Com. & Cha. & Full &Pri.
						&Full&Frontal
						&NME&FR\textsubscript{10\%}
						& NME & AUC\textsubscript{10\%} & FR\textsubscript{10\%}
						\\\hline\hline
						
						CPM+SBR\textsubscript{CVPR 18}~\cite{dong2018supervision}
						&3.28         & 7.58        & 4.10 & - 
						&2.14     & - 
						& - & -
						& - & - & - \\ 
						CPM+SBR\textsubscript{Fine-tuning}
						& $\textbf{3.24}_{\textcolor{red}{\downarrow1\%}}$		
						& \textbf{7.58}        
						& $\textbf{4.09}_{\textcolor{red}{\downarrow1\%}}$	
						& \textbf{5.27}
						& $\textbf{1.96}_{\textcolor{red}{\downarrow8\%}}$	
						& \textbf{1.77}
						&-&-
						& - & - & -
						\\\hline 
						
						LAB\textsubscript{CVPR 18}~\cite{wayne2018lab}
						& 2.98         & 5.19        & 3.49 &-  
						&    1.85       &     1.62    
						&    3.92       &      0.39
						& 5.27   & 53.23  & 7.56     \\
						LAB\textsubscript{Fine-tuning} 
						& $\textbf{2.91}_{\textcolor{red}{\downarrow3\%}}$        
						& $\textbf{5.16}_{\textcolor{red}{\downarrow1\%}}$         
						& $\textbf{3.35}_{\textcolor{red}{\downarrow4\%}}$  
						& \textbf{3.99}
						&$\textbf{1.84}_{\textcolor{red}{\downarrow1\%}}$ 
						&$\textbf{1.58}_{\textcolor{red}{\downarrow2\%}}$
						&$\textbf{3.69}_{\textcolor{red}{\downarrow5\%}}$
						&$\textbf{0.38}_{\textcolor{red}{\downarrow2\%}}$
						&$\textbf{5.11}_{\textcolor{red}{\downarrow3\%}}$ 
						&$\textbf{53.36}_{\textcolor{red}{\uparrow1\%}}$
						&$\textbf{7.40}_{\textcolor{red}{\downarrow2\%}}$
						\\\hline
						
						HRNetV2\textsubscript{CVPR 19}~\cite{sun2019high}   	& 2.87 				& \textbf{5.15}	 			& 3.32 & \textbf{3.85} &\textbf{1.57}&1.46    &    3.45   &      0.19
						& 4.60 & 56.42 & 4.84\\
						HRNetV2\textsubscript{Fine-tuning}
						& $\textbf{2.74}_{\textcolor{red}{\downarrow5\%}}$
						& 5.22 	 			
						& $\textbf{3.23}_{\textcolor{red}{\downarrow3\%}}$
						& 3.86
						&1.60
						&\textbf{1.46}
						&$\textbf{3.42}_{\textcolor{red}{\downarrow1\%}}$
						&$\textbf{0.18}_{\textcolor{red}{\downarrow3\%}}$ 
						&$\textbf{4.49}_{\textcolor{red}{\downarrow2\%}}$
						&$\textbf{56.71}_{\textcolor{red}{\uparrow1\%}}$
						&$\textbf{4.76}_{\textcolor{red}{\downarrow2\%}}$
						\\\hline
						AWing\textsubscript{ICCV 19}~\cite{wang2019adaptive}  & 2.72 				& 4.52 				& 3.07 	& 3.56& 1.53 & 1.38 & 4.94 & 0.99 
						& 4.36 & 57.19 & 2.84\\
						AWing\textsubscript{Fine-tuning} 
						& $\textbf{2.68}_{\textcolor{red}{\downarrow1\%}}$
						& $\textbf{4.50}_{\textcolor{red}{\downarrow1\%}}$
						& $\textbf{3.04}_{\textcolor{red}{\downarrow1\%}}$
						& $\textbf{3.54}_{\textcolor{red}{\downarrow3\%}}$
						&$\textbf{1.51}_{\textcolor{red}{\downarrow1\%}}$
						&$\textbf{1.29}_{\textcolor{red}{\downarrow6\%}}$
						&$\textbf{4.22}_{\textcolor{red}{\downarrow12\%}}$
						&$\textbf{0.85}_{\textcolor{red}{\downarrow14\%}}$
						&$\textbf{4.27}_{\textcolor{red}{\downarrow2\%}}$ 
						&$\textbf{57.48}_{\textcolor{red}{\uparrow1\%}}$
						&$\textbf{2.80}_{\textcolor{red}{\downarrow1\%}}$
						\\\hline
					\end{tabular}
				}
			\vspace*{0.1cm}
			\caption{
				Comparison between several fine-tuned models and the corresponding backbones on $4$ image datasets.
				Lower is better for NME and FR, while higher is better for AUC.
				Com., Cha., and Pri. in 300W dataset denote Common, Challenge, and Private subset, respectively.
			}
			\label{table:300W+AFLW+COFW}
			\vspace*{-0.1in}
		\end{table*}

	\subsection{Post-processing}
	\label{sec:post}
	
	The output of heatmap regression network is a stack of probability maps, the so-called heatmaps, which describe the opportunity of landmarks appearing at corresponding positions in images. The predicted landmark commonly locates at the centre of each resulted heatmap, which is obtained by selecting the maximum point (\textit{a.k.a.} \emph{argmax} method) in many previous works. A main drawback of the argmax method is the inability to achieve sub-pixel accuracy. Many previous works~\cite{bulat2017far, LuvizonPT18, tai2019towards} employ \emph{interpolation} methods to replace the argmax method for sub-pixel accuracy. Although these interpolation methods are demonstrated to achieve better performance on static images, their performance on videos is usually masked by random noise and motion blur. The reason lies in that the interpolation is sensitive to small errors on background pixels, which in turn increases the difficulty for estimating the heatmap center.
	
	Inspired by some related works on human pose estimation~\cite{chen2018cascaded, SunXWLW18}, we consider the significance of probability in discrete heatmaps and propose a Probability Density Centralization (\textbf{PDC}) \emph{Algorithm}~\ref{alg:algorithm} to compute the heatmap center to address the above problems. Contrary to the argmax or interpolation methods, PDC employs global information of the heatmaps to stabilize the predicted results. Moreover, we filter out the values below a certain threshold $\Theta_{PDC}$ to eliminate the interference by small errors on background pixels. The resulted center of mass is calculated by integrating the probability density of each pixel in the heatmap.
	
	 PDC is also designed as a replaceable post-processing method in the overall fine-tuning framework. Experiments on some public datasets prove that PDC as a post-processing algorithm can not only lift the accuracy over other methods on static images, but also make the predictions more stable in videos.
	
	\section{Experiments}
	\label{sec:experiments}
	\subsection{Evaluation Metrics}
	
	Previous related works generally utilize Normalized Mean Error (\textbf{NME}), Area-under-the-Curve (\textbf{AUC}), and Failure Rate (\textbf{FR}) for quantitative evaluations.
	However, these metrics cannot describe the stability of landmarks.
	In this section, we propose two novel metrics to evaluate the stability of landmarks for videos.
	These metrics are not mentioned in previous literatures on face alignment to the best of our knowledge.
	
	\paragraph{Mean Coefficient of Variation.}
	The coefficient of variation (CVar) is a statistical measure for the dispersibility of a group of data, which is given by the following equation:
	
	\begin{equation}
	CVar = \frac{\sqrt{\frac{1}{N-1}\sum_{i=1}^N (x_i - \overline{x})^2}}{\overline{x}},
	\end{equation}
	where $\mathit{N}$ denotes the number of samples and $\mathit{\overline{x}}$ denotes the average over all samples $x_i (i=1,2,...,N)$. 
	In the experiments, we use the mean coefficient of variation (\textbf{MCV}) to measure the average dispersibility of landmarks in the whole dataset. MCV is defined as follows:

	\begin{equation}
	MCV = \frac{\sum_{i=1}^{M} (CVar_i)}{M},
	\end{equation}
	where $\mathit{M}$ denotes the total number of videos and $\mathit{CVar_i}$ means the coefficient of variation for landmarks in each video.
	
	\paragraph{Mean Allan Variation.} In addition to the degree of data dispersion, data fluctuation~\cite{barnes1971characterization} is also an important indicator for data stability. Therefore, we introduce the Allan variance to evaluate the fluctuation of facial landmarks in the video, which is defined as follows:
	\begin{equation}
	AVar = \frac{\sum_{t=1}^{T-1} ({u}_{t+1} - {u}_{t})^2}{2(T-1)},
	\end{equation}
	where $\mathit{T}$ denotes the observation period and $\mathit{{u}_{t}}$ denotes the predicted landmark location of the $\mathit{t^{th}}$ frame.
	In the experiment of video datasets, we apply the mean Allan variation
	(\textbf{MAV}), which is averaged across all videos, to measure the average fluctuant frequency of landmarks.
	The MAV characterizes the cumulative inter-frame (local) stability, which is complementary to the global one as defined by MCV.

	\subsection{Implementation Details}\label{sec:details}
	The Adam optimizer~\cite{kingma2014adam} is used during the fine-tuning process, with an initial learing rate of $1e\!-\!4$ or $1e\!-\!5$.
	It also involves the PDC algorithm for the calculation of the Jitter loss function. The image data is not able to directly fed into training, since the proposed fine-tuning framework requires temporal information. We shift each original image and the corresponding landmarks by a few pixels, both horizontally and vertically, to construct a sequence with 5-8 frames. During testing, the ConvLSTM module is set to an initial state each time when the test data is entered to ensure fair comparisons.
	
	\subsection{Evaluation on Image Datasets}
	In order to make a fair comparison with previous works, we follow~\cite{sun2019high, ZhuLLT15} to split data and employ inter-ocular distance
	for face normalizations of 300W~\cite{sagonas2013300}, COFW~\cite{burgos2013robust}, and WFLW~\cite{wayne2018lab} dataset. 
	We use the provided face bounding box to normalize the AFLW~\cite{koestinger11a} dataset.
	We compute a bounding box that covers all labeled landmarks if no official boxes are provided. 
	For the WFLW dataset, a small number of the given boxes are too small to cover all landmarks,
	thus we enlarge them by $10\%$.
	The results of several SOTA methods on the four datasets are included in Table~\ref{table:300W+AFLW+COFW}, with both the baseline and the proposed fine-tuning methods in terms of quantitative metrics such as NME, AUC, and FR.
	
	
	\paragraph{Discussions on 300W and AFLW.}
	The proposed method gains notable improvement over the baseline methods with the fine-tuning framework.
	In particular, the NME is reduced by $5\%$ on the Common subsets of 300W and Frontal subset of AFLW.
	Furthermore, our method outperforms the SOTA methods on the Full set for both of the two datasets, which demonstrates the robustness of our approach against occlusion and large pose variations.
	There is an exception when using the HRNetV2 model as the baseline.
	This is caused by the negative values of heatmaps regressed by the vanilla HRNetV2, which is not fully compatible with our PDC post-processing method. We have tested that it can be avoided by changing the heatmap generation scheme.
	
	\begin{figure}[H]
		\vspace*{-0.1in}
		\centering
		\includegraphics[width=\linewidth]{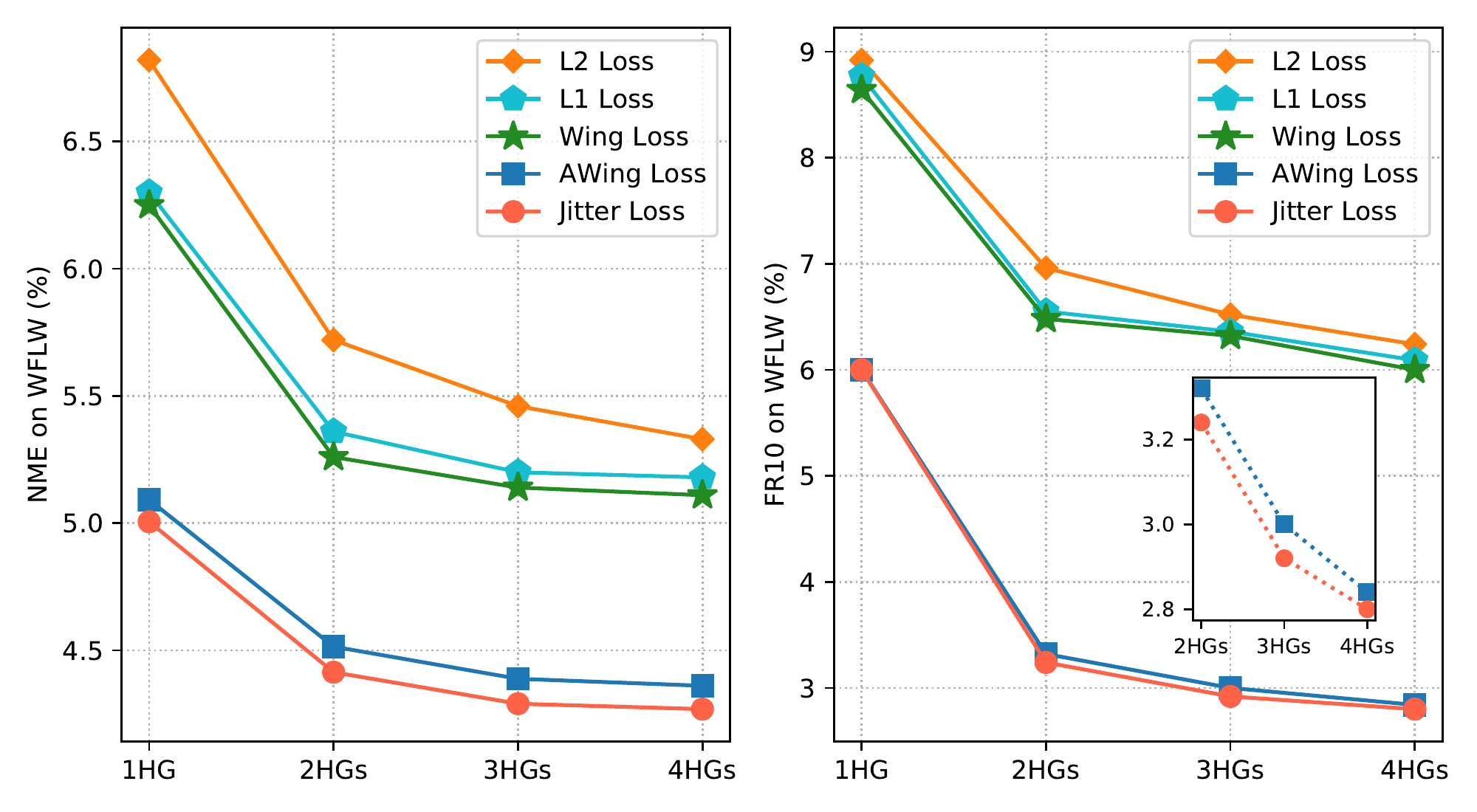}
		\vspace*{-0.4cm}
		\caption{
			Comparisons of different loss functions on the WFLW dataset when using the stacked HourGlass network as backbone.
		}
		\vspace*{-0.1in}
		\label{fig:Loss_NME_FR}
	\end{figure}
	
	\paragraph{Discussions on COFW and WFLW.}
	Experimental results show that the proposed method also gains notable improvement on both the COFW dataset and the WFLW dataset, in terms of quantitative metrics including NME, AUC, and FR.
	Specially, a tremendous improvement of over $12\%$ NME is achieved when using AWing method as baseline on the COFW dataset.
	In order to verify the proposed Jitter loss function, we conduct addition experiments on the most challenging WFLW dataset.
	We use the stacked HourGlass architectures~\cite{newell2016stacked} to compare the trained results by several different loss functions, as shown in Figure~\ref{fig:Loss_NME_FR}.
	Our proposed Jitter loss outperforms the existing ones in both NME and FR scores. This is a strong evidence demonstrating the effectiveness of the proposed loss function. 
	
	\begin{table*}
		\begin{center}
			\begin{tabular}{|c|p{1.1cm}p{1cm}p{1.1cm}|p{1.1cm}p{1cm}p{1.1cm}|p{1.1cm}p{1cm}p{1.1cm}|}
				\hline
				\multirow{2}{*}{Method} & \multicolumn{3}{c|}{Scenario 1} & \multicolumn{3}{c|}{Scenario 2} & \multicolumn{3}{c|}{Scenario 3} \\
				& NRMSE      & MCV     & MAV     & NRMSE      & MCV     & MAV     & NRMSE     & MCV     & MAV     \\ \hline\hline
				LAB\textsubscript{CVPR 18}~\cite{wayne2018lab}  &3.86&6.37&5.78	&3.53&6.72&7.00	&4.27&7.33&6.92\\
				LAB\textsubscript{Fine-tuning} 
				&$\textbf{3.07}_{\textcolor{red}{\downarrow20\%}}$
				&$\textbf{1.20}_{\textcolor{red}{\downarrow81\%}}$
				&$\textbf{2.36}_{\textcolor{red}{\downarrow59\%}}$
				&$\textbf{3.24}_{\textcolor{red}{\downarrow8\%}}$
				&$\textbf{1.95}_{\textcolor{red}{\downarrow71\%}}$
				&$\textbf{2.36}_{\textcolor{red}{\downarrow66\%}}$
				&$\textbf{3.88}_{\textcolor{red}{\downarrow9\%}}$
				&$\textbf{2.46}_{\textcolor{red}{\downarrow65\%}}$
				&$\textbf{3.59}_{\textcolor{red}{\downarrow48\%}}$
				\\\hline
				
				HRNetV2\textsubscript{CVPR 19}~\cite{sun2019high}&2.94&5.62&4.23	&2.96&6.95&6.92		&2.97&7.52&7.37\\
				HRNetV2\textsubscript{Fine-tuning} 
				&$\textbf{2.52}_{\textcolor{red}{\downarrow14\%}}$
				&$\textbf{1.44}_{\textcolor{red}{\downarrow74\%}}$
				&$\textbf{1.61}_{\textcolor{red}{\downarrow62\%}}$
				&$\textbf{2.68}_{\textcolor{red}{\downarrow10\%}}$
				&$\textbf{1.75}_{\textcolor{red}{\downarrow75\%}}$
				&$\textbf{1.82}_{\textcolor{red}{\downarrow74\%}}$
				&$\textbf{2.88}_{\textcolor{red}{\downarrow3\%}}$
				&$\textbf{2.06}_{\textcolor{red}{\downarrow73\%}}$
				&$\textbf{2.98}_{\textcolor{red}{\downarrow60\%}}$
				\\\hline
				
				AWing\textsubscript{ICCV 19}~\cite{wang2019adaptive}  &2.77&5.79&6.30	&2.94&6.13&7.17	&3.02&6.01&7.20\\
				AWing\textsubscript{Fine-tuning}
				&$\textbf{2.49}_{\textcolor{red}{\downarrow10\%}}$
				&$\textbf{1.09}_{\textcolor{red}{\downarrow81\%}}$
				&$\textbf{1.57}_{\textcolor{red}{\downarrow74\%}}$
				&$\textbf{2.65}_{\textcolor{red}{\downarrow10\%}}$
				&$\textbf{1.13}_{\textcolor{red}{\downarrow82\%}}$
				&$\textbf{1.91}_{\textcolor{red}{\downarrow73\%}}$
				&$\textbf{2.92}_{\textcolor{red}{\downarrow4\%}}$
				&$\textbf{1.33}_{\textcolor{red}{\downarrow78\%}}$
				&$\textbf{3.04}_{\textcolor{red}{\downarrow58\%}}$
				\\\hline
				
				\multicolumn{10}{|c|}{Comparisons with SOTA video-oriented methods}
				\\\hline		
				CPM+SBR\textsubscript{CVPR 18}~\cite{dong2018supervision}&5.37&3.62&3.54	&4.55&3.99&3.23		&5.92&4.22&3.11\\
				FHR+STA\textsubscript{AAAI 19}~\cite{tai2019towards}&3.93&3.01&2.33	&3.82&3.15&2.87		&4.91&3.36&3.27\\
				\hline
				Best of Ours\textsubscript{w/o PDC}   &2.52&1.44&1.61   &2.68&1.75
				&$\textbf{1.82}_{\textcolor{red}{\downarrow36\%}}$
				&$\textbf{2.88}_{\textcolor{red}{\downarrow41\%}}$
				&2.06
				&$\textbf{2.98}_{\textcolor{red}{\downarrow9\%}}$\\
				Best of Ours\textsubscript{with PDC}
				&$\textbf{2.49}_{\textcolor{red}{\downarrow37\%}}$
				&$\textbf{1.09}_{\textcolor{red}{\downarrow63\%}}$
				&$\textbf{1.57}_{\textcolor{red}{\downarrow33\%}}$
				&$\textbf{2.65}_{\textcolor{red}{\downarrow31\%}}$
				&$\textbf{1.13}_{\textcolor{red}{\downarrow58\%}}$
				&1.91
				&2.92
				&$\textbf{1.33}_{\textcolor{red}{\downarrow60\%}}$
				&3.04\\
				\hline
			\end{tabular}
		\end{center}
	    \vspace*{-0.1in}
		\caption{
			Comparisons of NRMSE, MCV, and MAV on 300VW dataset. Lower is better.
			The results show that our fine-tuning pipeline can improve the stability and accuracy of the backbone models.
			Moreover, our method also achieves superior results over the SOTA video-oriented methods (SBR and STA).
		}
		\label{table:300VW}
		\vspace*{-0.1in}
	\end{table*}
	
	\subsection{Evaluation on Video Datasets \protect\footnote{Some examples of comparative videos can be found at \url{https://www.bilibili.com/video/BV1Pa4y1J7tU}}}
	\paragraph{300VW.}
	We follow~\cite{keqiang2019fab} to use 50 videos for training and the rest 64 videos for testing on the 300VW dataset~\cite{shen2015first}.
	The test set is divided into well-lit (Scenario 1), mild unconstrained (Scenario 2), and challenging (Scenario 3) categories according to the difficulties.
	Normalized Root Mean Squared Error (\textbf{NRMSE}) is employed as the accuracy evaluation metric, 
	and we also apply the MCA and MAV for stability evaluation.
	We show the comparative results with several SOTA methods including both image-oriented and video-oriented ones in Table~\ref{table:300VW}. 
	Our proposed fine-turning framework can reduce the NRMSE significantly by $14\%$ on average. 
	Moreover, the MCV and MAV are reduced by at least $40\%$ compared to the video-oriented methods. 
	The results achieved by the PDC post-processing methods is on par with but no better than that without PDC for the specific Scenario $3$, which may be caused by noises from global heatmap regression.
	
	\begin{figure}[H]
		\vspace*{-0.1in}
		\begin{center}
			\includegraphics[width=\linewidth]{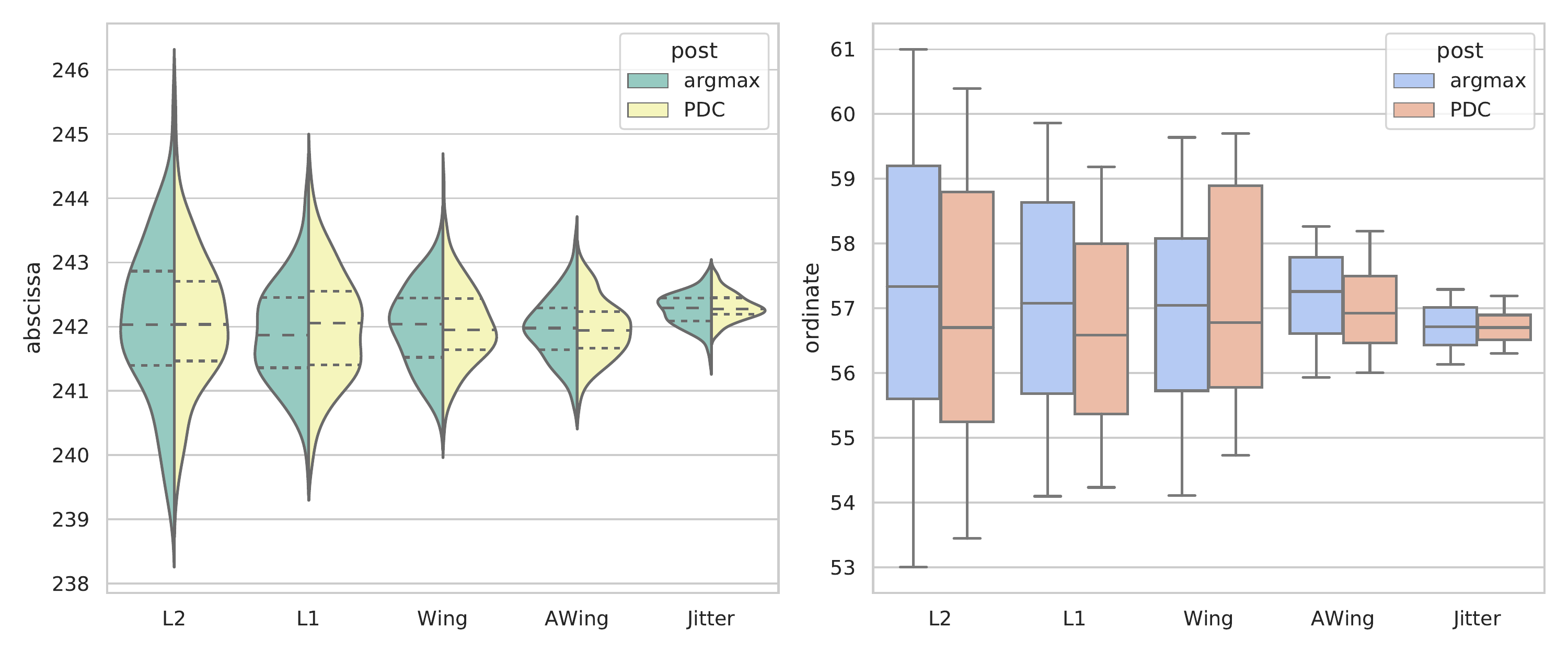}
		\end{center}
		\vspace*{-0.1in}
		\caption{
			Comparisons of several loss functions with two different post-processing methods on the unlabeled DDD dataset. 
		}
		\vspace*{-0.1in}
		\label{fig:post_processing_functions}
	\end{figure}
	
	\paragraph{Unlabeled Raw Dataset.}
	Manual labeling of landmarks usually suffers from inevitable errors.
	In our experiment for assessing the PDC method on the 300VW dataset, the labeling errors, together with the notable face movement on videos, hinder the results of different post-processing methods.
	Therefore, we carry out our study on an unlabeled Driver Drowsiness Detection (DDD) dataset~\cite{weng2016driver}, which includes more than 100 videos with small face motions.
	Comparisons of both different loss functions and post-processing methods are shown in Figure~\ref{fig:post_processing_functions}.
	The violin diagram on the left figure reveals the Jitter loss enables the predicted abscissa more compact than others, and this is the same to the right box diagram for the ordinate.
	Overall the Jitter loss shrinks the distribution of predicted landmarks by at least $50$\% compared with other loss functions.  
	Moreover, this experiment also illustrates that the PDC algorithm contributes to some improvements as opposed to the argmax method.
	
	\begin{figure}[H]
	    \vspace*{-0.1in}
		\centering
		\includegraphics[width=\linewidth]{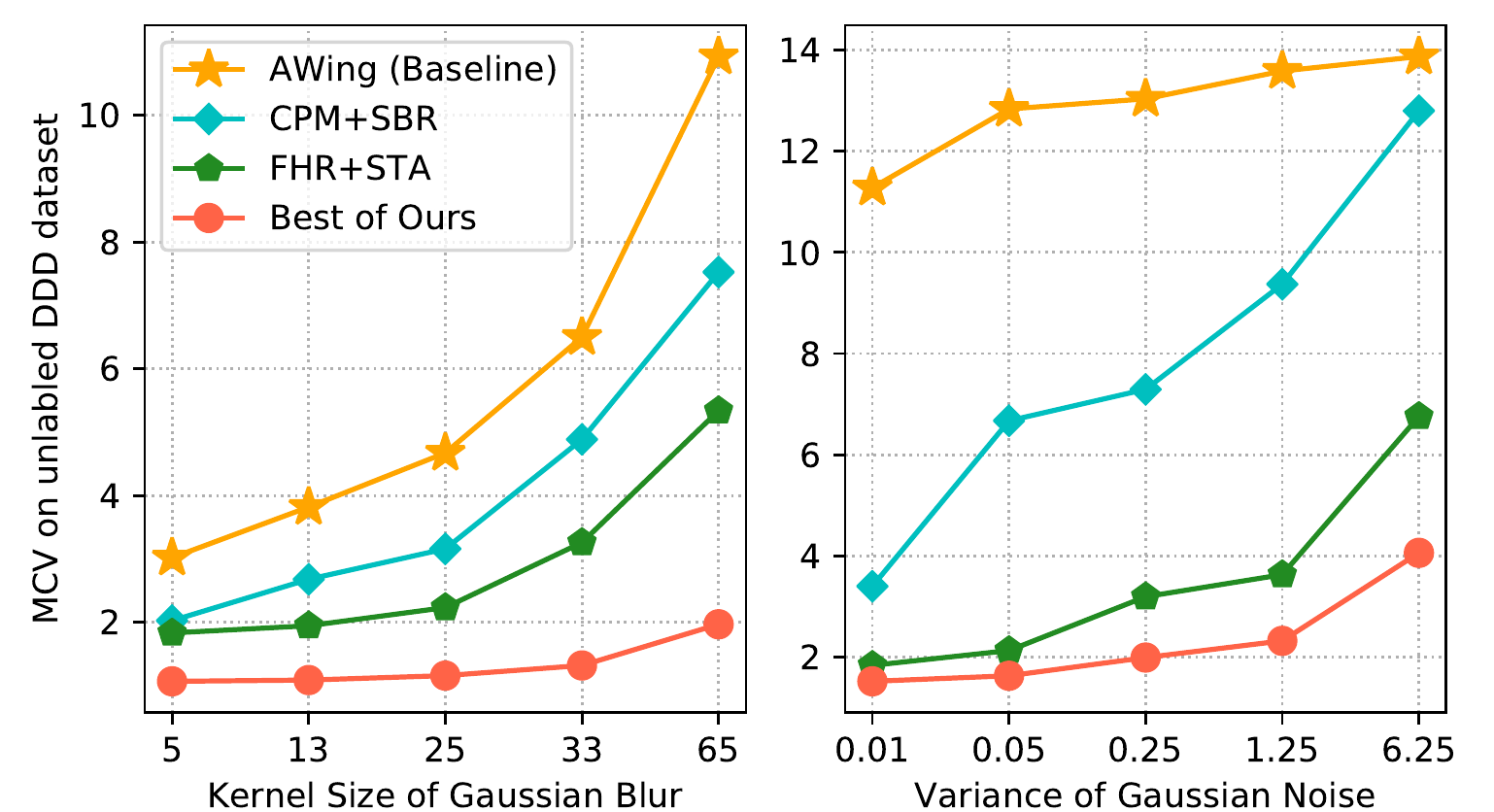}
		\vspace*{-0.1in}
		\caption{
			Comparisons of the stability of predictions on videos with random Gaussian noise and blurs.
			The proposed framework achieves robust predictions with significantly lower MCV owing to the effective use of temporal information.
		}
		\vspace*{-0.1in}
		\label{fig:blur_noise_mcv}
	\end{figure}

	\paragraph{Noise and Blurring Effects on Videos.}
	Contrary to still images, the real-life videos are usually contaminated by various noise and motion blurs. We conduct experiments on the unlabeled DDD dataset to test the proposed method on its robustness to noise and motion blur.
	We add different levels of simulated Gaussian noise and blurring effects with Gaussian kernels to $30$ selected segments of videos with small face movements. 
	The comparative results with several SOTA methods on videos are shown in Figure~\ref{fig:blur_noise_mcv}, where the image-oriented AWing loss is employed as a baseline method.
	Obviously, the video-oriented methods outperform the image-oriented one in terms of MCV in the noisy cases, while our proposed method suppresses other methods by a clear margin, especially in extreme cases.
	
	\begin{table}
		\begin{center}
			\begin{tabular}{|c|ccccc|}
				\hline
				$\Theta$ 		   & 0 & 0.5 & 1 & 1.5 & 2 \\ \hline\hline
				HRNetV2            & 10.49 & 5.24 & \textbf{3.88} & 5.26 & 9.09 \\
				LAB(4HGs)          & 13.23  & 11.46  & \textbf{6.21}  & 9.46  & 12.57   \\
				CPM                & 29.84  & 18.22  & 16.82  & \textbf{16.47}  & 17.29   \\
				\hline
			\end{tabular}
		\end{center}
		\caption{
			Evaluation on different hyperparameter settings of Jitter loss via NRMSE*MCV metric on 300VW dataset (Scenario 3).
		}
		\label{table:search}
		\vspace*{-0.1in}
	\end{table}

	\subsection{Hyperparameters Selection}
	Two hyperparameters remains to be set are the thresholds $\Theta$ for the Jitter loss and $\Theta_{PDC}$ to filter out small values for the PDC algorithm.
	We adopt several backbone networks to evaluate different $\Theta$ on 300VW dataset. In principle, there is a trade-off between the evaluation metrics of MCV and NRMSE. More specifically, the MCV should decrease as $\Theta$ increases, while the NRMSE should increase with $\Theta$.
	However, since the backbone networks have already achieved low NRMSEs, a small $\Theta$ is enough to improve detection stability significantly.
	We show the results in terms of the product of MCV and NRMSE for different backbone networks in Table~\ref{table:search}, where the optimal $\Theta$ varies within a small range. Generally the optimal $\Theta$ increases with the prediction error of the backbone network.
	Table~\ref{table:search} also indicates that the proposed method is mildly sensitive to $\Theta$, thus we set $\Theta=1$ in this paper.
	
	In addition, we test the influence of $\Theta_{PDC}$ while fixing $\Theta$ to be optimal. Table~\ref{table:different_theta} shows the results when using the HRNetV2 as backbone network. On the one hand, increasing $\Theta_{PDC}$ can effectively smooth the small fluctuations of landmarks as well as maintaining the accuracy by neglecting small error on background pixels. On the other
	hand, an excessively large threshold will wipe out useful  information and lead to performance degradation.
	Therefore, a suitable $\Theta_{PDC}$ is crucial for the fine-tuning framework over the backbone network.
	We set $\Theta_{PDC}=0.2$ in our experiments according to the result in Table~\ref{table:different_theta}.

	\begin{table}
		\begin{center}
			\begin{tabular}{|c|ccccc|}
				\hline
				$\Theta_{PDC}$  & 0 & 0.1 & 0.2 & 0.4 & 0.6 \\ \hline\hline
				NRMSE           & 3.19   & \textbf{2.92}  & 2.94   & 3.05   & 2.99 \\ 
				MAV             & 3.26   & 3.11  & \textbf{3.04}  & 3.07   & 3.10 \\ 
				NRMSE*MAV       & 10.40  & 9.08  & \textbf{8.94}  & 9.36   & 9.26 \\
				\hline
			\end{tabular}
		\end{center}
		\caption{
			Evaluation on different values of $\Theta_{PDC}$ when using pre-trained HRNetV2 as backbone.
			The results demonstrate that an appropriate threshold can improve both accuracy and stability.}
		\label{table:different_theta}
		\vspace*{-0.1in}
	\end{table}
	
	\section{Conclusions}
	In this paper, we propose a novel fine-tuning framework for stable face alignment. The proposed framework is capable of converting state-of-the-art facial landmark detector for images to that for videos. The core of our framework is the ConvLSTM structure with the Jitter loss for fine-tuning, which provides easy access to leverage temporal information to suppress inaccurate and jittered landmarks during training. Extensive experiments demonstrate the effectiveness of the proposed method and its superior performance over the existing ones. Moreover, it provides a general framework to improve both the accuracy and stability by taking advantage of the recent progress in face alignment without retraining the entire network.
	
	{\small
		\bibliographystyle{ieee_fullname}
		\bibliography{egbib}
	}
	
\end{document}